\documentclass[letterpaper, 10 pt, conference]{ieeeconf}
\IEEEoverridecommandlockouts
\usepackage{cite}
\usepackage{lmodern}
\usepackage{graphicx}
\usepackage{times}
\usepackage{ragged2e}
\usepackage{amsmath,amssymb, amsfonts}
\usepackage{tabularx}
\usepackage{hyperref}
\usepackage{cleveref}
\usepackage{siunitx}
\usepackage{dblfloatfix}
\usepackage{authblk}
\usepackage[table]{xcolor}
\usepackage[T1]{fontenc}
\usepackage{siunitx}
\usepackage{booktabs}
\usepackage[caption=false]{subfig}
\usepackage{algorithm}
\usepackage{algpseudocode}

\begin{document}
\noindent

\title{\LARGE \bf
The Hybrid Extended Bicycle: A Simple Model for High Dynamic Vehicle Trajectory Planning
}

\author{
Agapius Bou Ghosn$^{1}$,
Philip Polack$^{1}$,
and Arnaud de La Fortelle$^{1,2}$
\thanks{$^{1}$ Center for Robotics, Mines Paris, PSL University, 75006 Paris, France {\tt [agapius.bou\textunderscore ghosn, philip.polack, arnaud.de\textunderscore la\textunderscore fortelle]@minesparis.psl.eu}}
\thanks{$^{2}$ Heex Technologies, Paris, France}
}

\maketitle

\thispagestyle{empty}
\pagestyle{empty}

\begin{abstract}
While highly automated driving relies most of the time on a smooth driving assumption, the possibility of a vehicle performing harsh maneuvers with high dynamic driving to face unexpected events is very likely. The modeling of the behavior of the vehicle in these events is crucial to proper planning and controlling; the used model should present accurate and computationally efficient properties to ensure consistency with the dynamics of the vehicle and to be employed in real-time systems. In this article, we propose an LSTM-based hybrid extended bicycle model able to present an accurate description of the state of the vehicle for both normal and aggressive situations. The introduced model is used in a Model Predictive Path Integral (MPPI) plan and control framework for performing trajectories in high-dynamic scenarios. The proposed model and framework prove their ability to plan feasible trajectories ensuring an accurate vehicle behavior even at the limits of handling. 
\end{abstract}

\section{Introduction}
The ability to operate a vehicle in emergency scenarios is essential for its safety. Aggressive maneuvers are not highly encountered but can pose threats to the safety of a vehicle not prepared to deal with them. An autonomous vehicle is expected to have the ability to plan feasible trajectories and to apply controls to follow the planned trajectories in all scenarios. 

Trajectory planning for autonomous vehicles makes use of three main categories of planners \cite{gonzalez_review_2016}, graph search based planners such as A* planners (e.g. \cite{yoon_recursive_2015}), sampling based planners such as Rapidly-exploring Random Trees (RRT) planners (e.g. \cite{feraco_local_2020}), or optimal control based planners such as Model Predictive Control (MPC) planners which have been widely used in autonomous vehicle applications in the recent years (e.g. \cite{ji_path_2017}, \cite{dixit_trajectory_2020}). All of the mentioned planners depend on a vehicle model that is required to accurately represent the behavior of the vehicle. The accuracy of the used vehicle model directly affects the vehicle's actions, thus its safety.

\begin{figure}[htp]

\subfloat[Harsh lane change maneuver using the proposed hybrid extended bicycle based planner.]{%
  \includegraphics[width=\columnwidth]{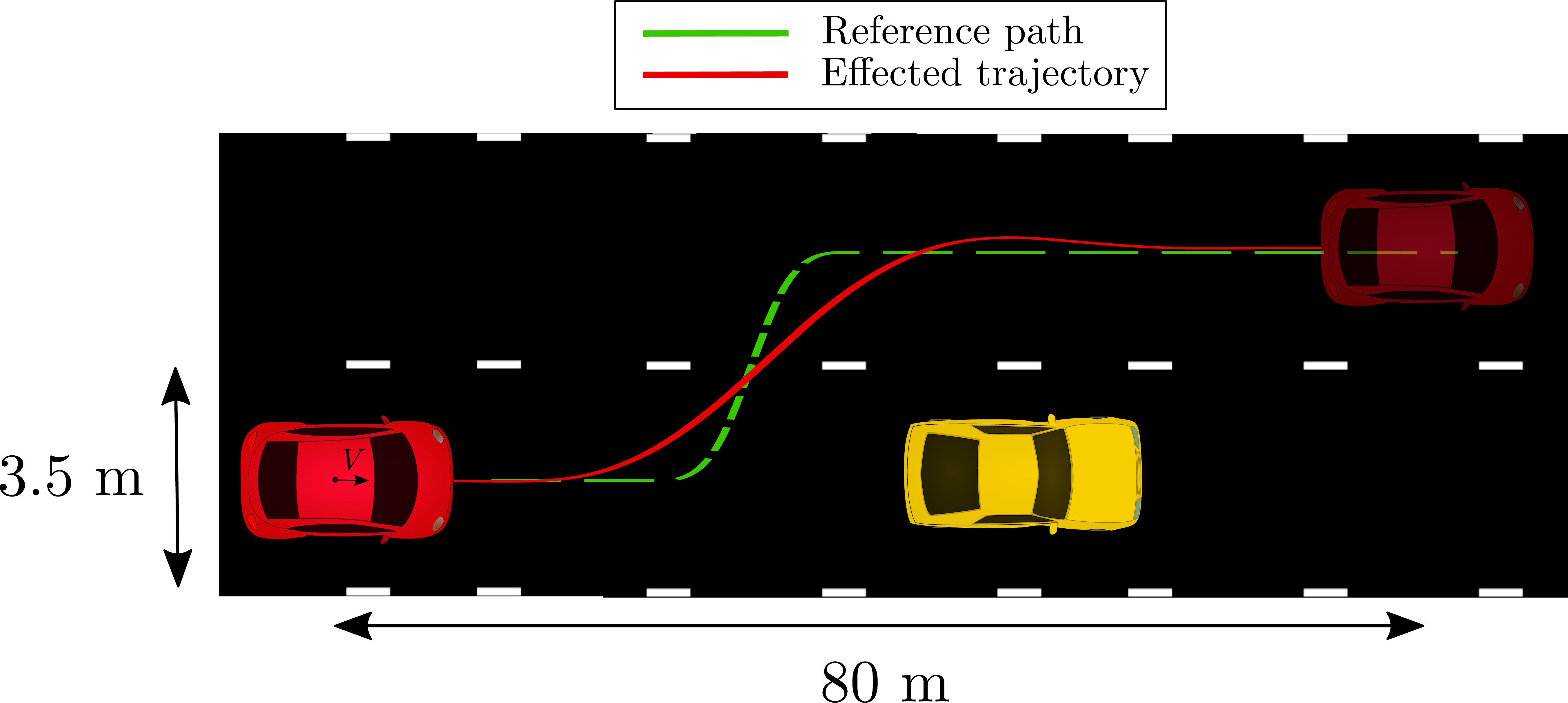}%
}

\subfloat[Harsh lane change maneuver using the kinematic bicycle based planner.]{%
  \includegraphics[width=\columnwidth]{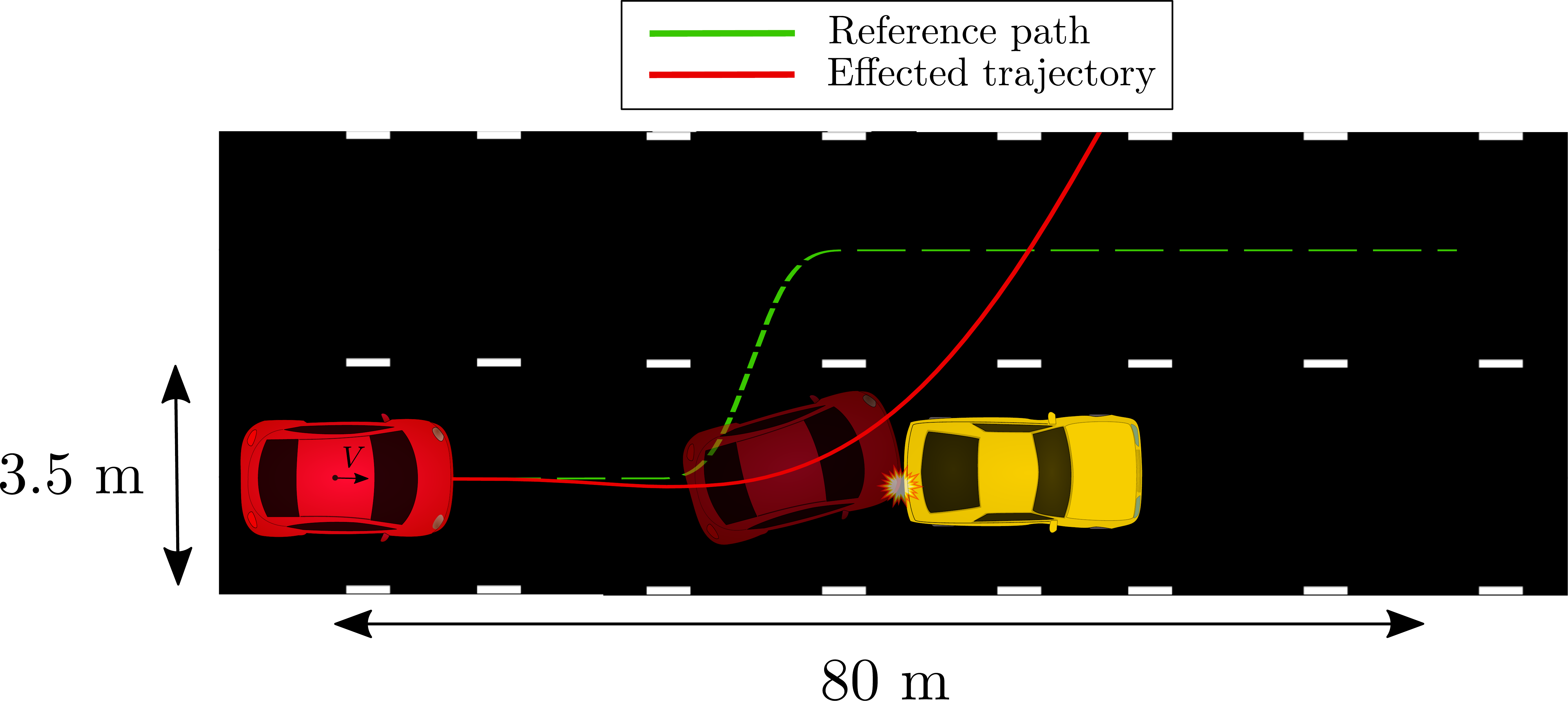}%
}

\caption{Comparison between the vehicle behavior using the proposed hybrid extended bicycle planner and a kinematic bicycle planner to effect a harsh lane change maneuver ($V_{\text{ref}}=\SI{25}{\meter\per\second}$). The proposed method performs a more accurate maneuver with lateral accelerations reaching $a_y^\text{max} = 0.75\text{g}$. \label{laneChange.fig}}

\end{figure}

The difficulty of predicting the vehicle behavior in high dynamics relies in the resulting high nonlinearities (especially at the tire level) which defy the assumptions used for normal driving. In these scenarios, assumptions related to linear tire behavior, or no-slip conditions that are usually employed in simple models are not valid anymore. This urges designers to upgrade to more complex vehicle and tire models. While this guarantees a larger validity domain, it creates two problems: on the one hand, using complex models introduces computational deficiency to the planner that is supposed to make fast decisions; on the other hand, the usage of additional parameters in complex models may put the robustness of these models in jeopardy. 

The literature includes many works that use a kinematic bicycle model to solve the motion planning problem of a vehicle (e.g. \cite{cardoso_model-predictive_2017}, \cite{zhang_optimal_2020}, \cite{wang_path_2022}). While the kinematic bicycle model is a simple model with computationally efficient properties, its validity domain is limited to lateral accelerations $a_y<0.5\mu g$ as proven in \cite{polack_kinematic_2017}.  Thus, using the kinematic bicycle model to plan trajectories may result in infeasible ones, especially when in high dynamics, i.e. high lateral accelerations. 

Other works make use of more complex models as the dynamic bicycle model used in \cite{feraco_local_2020} with a linear tire model or in \cite{li_dynamic_2019} with a Pacejka \cite{pacejka_tyre_2006} tire model. In the first work, the operational range of the planner is limited to the linear region of the tire model, i.e. low dynamic maneuvers; while in the second work, the use of the Pacejka model requires the knowledge of many parameters that are not easily accessible. In addition, implementing tire dynamics imposes the usage of lower integration time steps due to the fast dynamics of the wheels limiting the planner to lower planning horizons.

This article targets the problem of representing the behavior of the vehicle in high dynamics using a simple model to plan feasible trajectories.

We present a hybrid vehicle model that describes the motion of the vehicle in low and high dynamic maneuvers. The presented method makes use of the extended bicycle model \cite{lenain_adaptive_2003}, an augmentation of the kinematic bicycle model. We integrate recurrent neural networks in the extended model to correct the modeling errors and account for slipping angles at the wheels. In summary, this paper presents the following contributions:
\begin{itemize}
    \item We augment the extended bicycle model to a hybrid model that uses recurrent neural networks to represent the motion of the vehicle even at the limits of handling. 
    \item We present a planning architecture based on the Model Predictive Path Integral (MPPI) \cite{williams_aggressive_2016} that uses the developed hybrid model to plan trajectories in different scenarios.
    \item We compare the proposed planner to a kinematic bicycle based planner knowing that the kinematic model is heavily used in the literature for planning purposes (check Fig. \ref{laneChange.fig}). 
\end{itemize}

In the rest of this paper, the reference model is a four-wheel vehicle model with a Pacejka tire model simulated by \cite{polack_consistency_2018} and used as a reference model in several works (e.g. \cite{polack_kinematic_2017}, \cite{qian_optimal_2016}, \cite{devineau_coupled_2018}).  The paper will start by presenting the reference model in Section \ref{referenceModel.sec}, both the kinematic bicycle model and the extended bicycle model are then presented in Sections \ref{KBM.sec} and \ref{EBM.sec} respectively. The proposed approach, including the data generation procedure, the network architecture and training, and the planner are detailed in Sections \ref{proposedApproach.sec} and \ref{planner.sec}. Testing of the proposed approach is then presented in Section \ref{results.sec} with a comparison to a kinematic bicycle based planner. The paper is concluded in Section \ref{conclusion.sec}.

\section{Models}
In this section we present the different models to be used for our development and analysis. As stated before, the reference model is a four-wheel vehicle model defined in \cite{polack_consistency_2018}. The kinematic bicycle model and the extended bicycle model are both simplified vehicle models. The extended model will be the basis of the proposed approach while the kinematic bicycle model will be used for comparison purposes only. The reference model will be briefly presented in Section \ref{referenceModel.sec}, followed by the kinematic bicycle model in Section \ref{KBM.sec} and the extended bicycle model in Section \ref{EBM.sec}.

\subsection{Reference Model} \label{referenceModel.sec}
\begin{figure}
    \centering
    \includegraphics[width=.8\columnwidth]{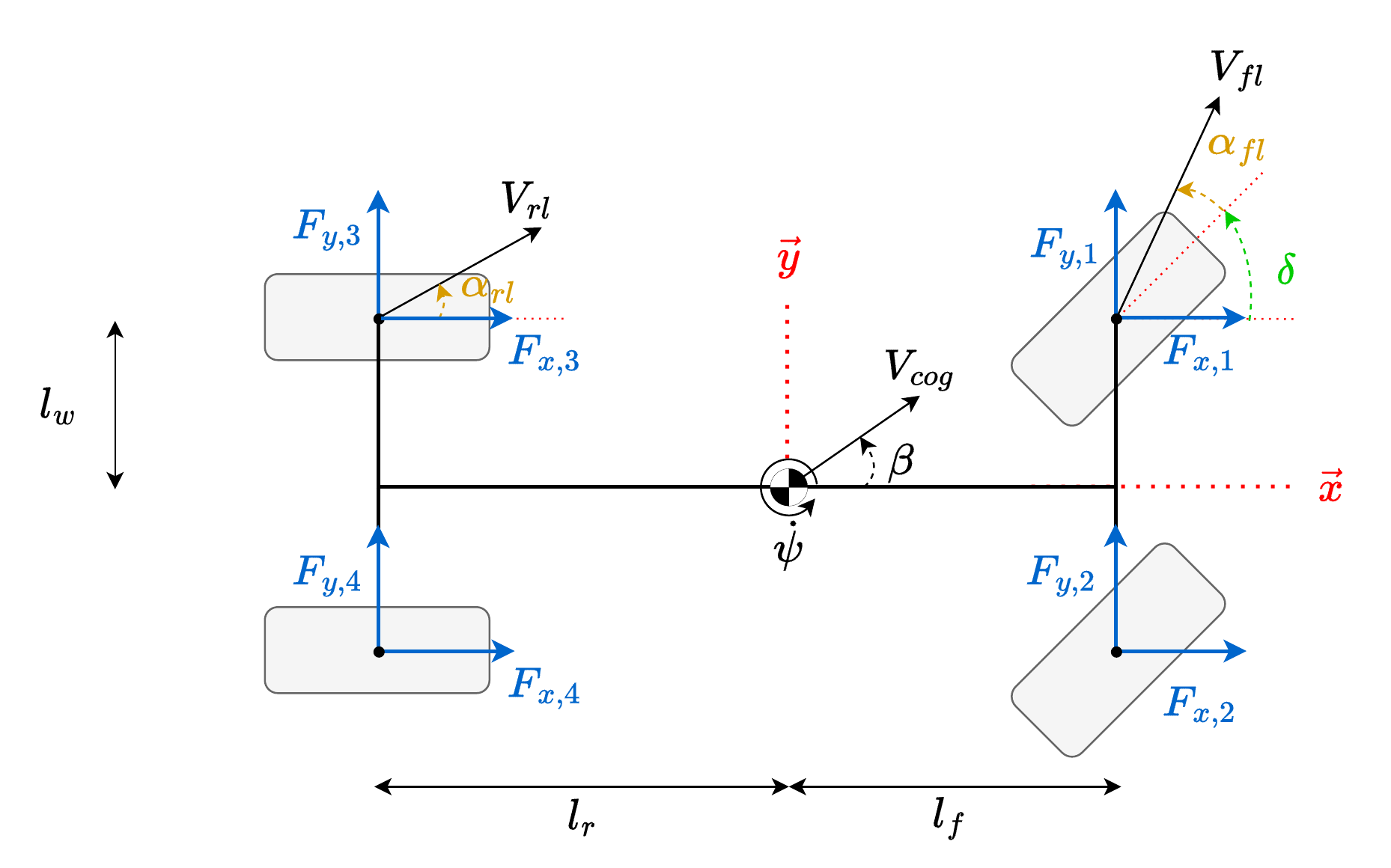}
    \caption{The four-wheel vehicle model.}
    \label{fourWheelModel.fig}
\end{figure}
The reference model used to describe the dynamics of the vehicle is a 9 degrees of freedom (DoF) four-wheel dynamic model with a Pacejka tire model \cite{pacejka_tyre_2006}. It can accurately describe the motion of the vehicle and it was chosen as a reference in several works as mentioned in the previous section. The states, parameters and control inputs describing the model are presented in Table \ref{refModelParameters.tab}. As illustrated in Fig. \ref{fourWheelModel.fig}, the model is made of two front steerable wheels and two rear non-steerable wheels. 
The motion of the model is represented by the velocity vector issued from the center of gravity ($cog$) of the vehicle. 
Furthermore, it is assumed that the roll and pitch occur around the center of gravity and that the aerodynamic forces are represented by a single force at the vehicle's front. The model considers the coupling of longitudinal and lateral slips and the load transfer between tires.

\begin{table}
\centering
\caption{States, parameters and control inputs for the reference \\9 DoF model}
    \begin{tabular}{cp{.7\columnwidth}c}
    \toprule
    \textbf{States} & \textbf{Description} \\
    \midrule
	$X$, $Y$ & Coordinates of the center of gravity in the inertial frame \\
    $\theta$, $\phi$, $\psi$ & Roll, pitch and yaw angles of the vehicle body \\
    $\omega_i$ & Angular speed of the wheel $i$\\
    $V_x$, $V_y$ & Longitudinal and lateral speeds in the vehicle's frame\\
	$\dot\theta$, $\dot\phi$, $\dot\psi$ & Roll, pitch and yaw angular velocities\\
    \midrule
    \textbf{Parameters} & \textbf{Description} \\
    \midrule
    $I_x$, $I_y$, $I_z$ & Inertia of the vehicle around its roll, pitch and yaw axes\\
    $F_{xpi}$, $F_{ypi}$ & Longitudinal and lateral tire forces in the tire frame\\
    $F_{xi}$, $F_{yi}$ & Longitudinal and lateral tire forces in the vehicle frame\\
    $F_{zi}$ & Normal reaction force on wheel $i$\\
    $M$ & Mass of the vehicle \\
    $l_f$, $l_r$ & Distance between the front (or rear) axle and the center of gravity\\
    $l_w$ & Half-width of the vehicle\\
	$h$ & Height of the center of gravity\\
    $r_{\text{eff}}$ & Effective wheel radius \\
    \midrule
    \textbf{Control inputs} & \textbf{Description} \\
    \midrule
    $T_{w,i}$ & Torque at each wheel\\
    $\delta$ & Steering angle of the front wheels\\ 
    \midrule
    \end{tabular}
\label{refModelParameters.tab}
\end{table}

The state of the model is defined as 
$
\begingroup
\setlength\arraycolsep{4.25pt}
Z = \left[\begin{matrix}X & V_x & Y & V_y & \psi & \dot\psi & \theta & \dot\theta & \phi  \end{matrix}\begin{matrix}& \dot\phi
&\omega_1 & \omega_2 & \omega_3 & \omega_4\end{matrix}\right]^T
\endgroup
$
The equations governing the state evolution of the model can be found at the reference. 

The presented model serves as the reference vehicle for this paper. It will be used to generate data points to train the proposed approach at a first stage, and will be controlled based on the planned trajectories at a second stage. Planned trajectories will be computed based on the kinematic and extended bicycle models presented next. 

\subsection{Kinematic Bicycle Model}\label{KBM.sec}
\begin{figure}
    \centering
    \includegraphics[width=.9\columnwidth]{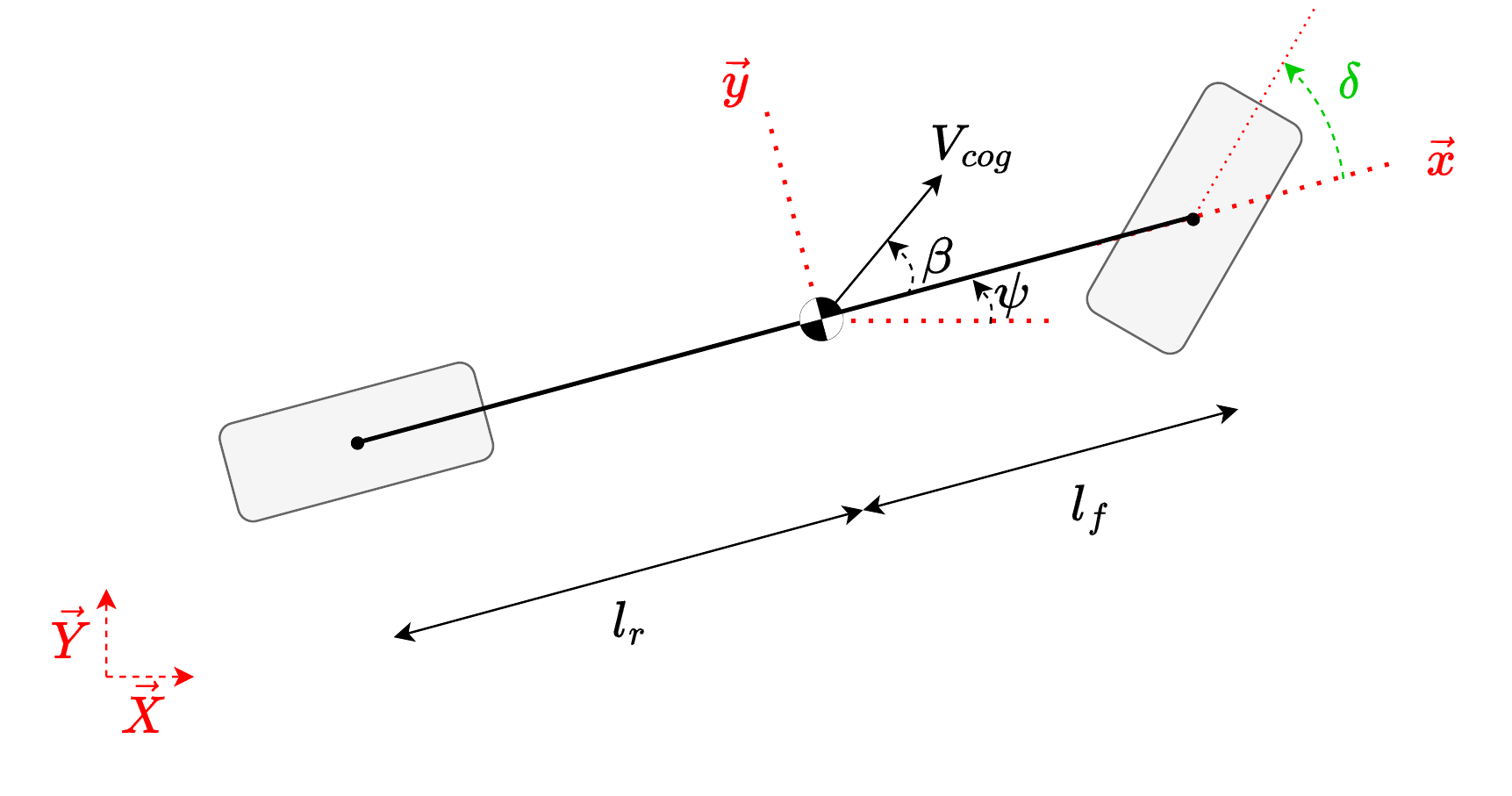}
    \caption{The kinematic bicycle model.}
    \label{kinematicModel.fig}
\end{figure}
The kinematics study of a system is concerned with the motion of the system without reference to the forces or masses entailed in it.
Several assumptions are added to the 9 DoF model to reach the kinematic bicycle model:
\begin{enumerate}
\item The four-wheel model is simplified into a bicycle model: front wheels are considered as a single steerable wheel, rear wheels as a single non-steerable wheel;
\item The pitch and roll dynamics are neglected;
\item The slip angles at both wheels are neglected.
\end{enumerate}
The model, with a center of gravity reference, is illustrated in Fig. \ref{kinematicModel.fig} and its states,  parameters and inputs are presented in Table \ref{bicycleParameters.tab}.

\begin{table}
\centering
\caption{States, parameters and control inputs for the bicycle model.}
    \begin{tabular}{cp{.7\columnwidth}c}
      \toprule
      \textbf{States} & \textbf{Description} \\
      \midrule
      $X$, $Y$ & Coordinates of the center of gravity in the inertial frame \\
      $\psi$ & Yaw angle \\
      $\beta$ & Side slip angle at the center of gravity\\
      \midrule
      \textbf{Parameters} & \textbf{Description} \\
      \midrule
      $l_f$, $l_r$ & Distance between the front (or rear) axle and the center of gravity\\
      \midrule
      \textbf{Control inputs} & \textbf{Description} \\
      \midrule
      $V$ & Speed at the center of gravity \\
      $\delta$ & Steering angle of the front wheel\\
      \bottomrule
    \end{tabular}
\label{bicycleParameters.tab}
\end{table}

The state of the model is defined as 
$
Z =  \begin{bmatrix}X & Y & \psi \end{bmatrix}^T \nonumber
$
and its evolution is defined as follows:
\begin{subequations}
\begin{eqnarray}
\dot X &=& V\cos(\psi+\beta) \\
\dot Y &=& V\sin(\psi+\beta) \\
\dot\psi &=& \frac{V\tan\delta\cos\beta}{l_f+l_r} \\
\beta &=& \arctan\left(\frac{l_r \tan\delta}{l_f+l_r}\right) 
\end{eqnarray}\label{kmodel.eq}
\end{subequations}

As mentioned earlier the kinematic bicycle model is widely used for planning purposes in the literature; for this reason we will compare the performance of the proposed approach to it. A more accurate but simple model is presented next; it is an extension to the kinematic bicycle model. 

\subsection{Extended Bicycle Model}\label{EBM.sec}
\begin{figure}
    \centering
    \includegraphics[width=.9\columnwidth]{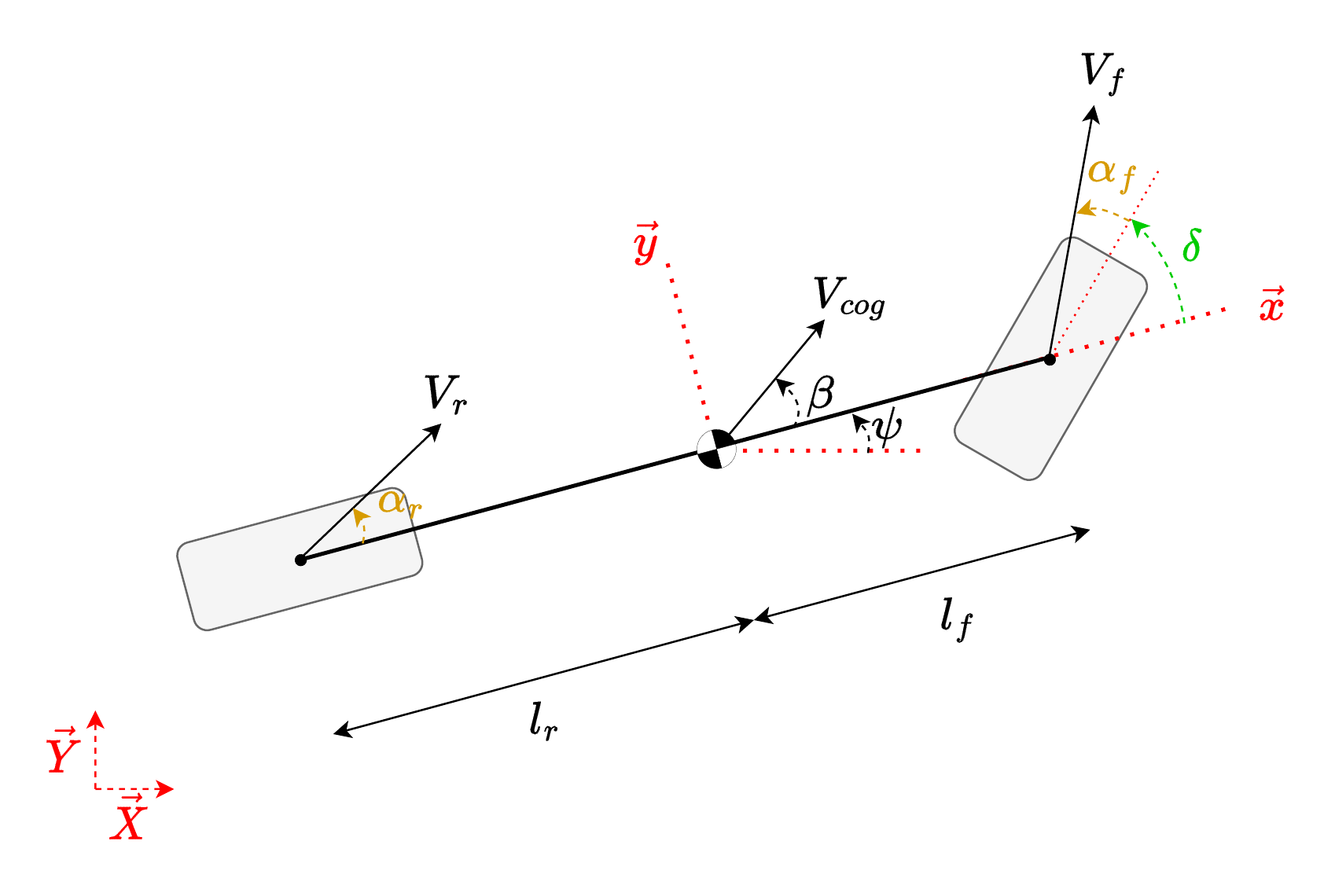}
    \caption{The extended bicycle model.}
    \label{extendedModel.fig}
\end{figure}
The extended bicycle model (EBM) shown in Fig. \ref{extendedModel.fig} does not take into account the dynamics of the system as well. It is concerned with the kinematics of the vehicle while including the slip angles at the front and rear wheels. This model was introduced in \cite{lenain_adaptive_2003} as an augmentation of the kinematic bicycle model presented previously to account for slips in the vehicle's motion. 
As it is the case for the previous model, the state of the extended model taking as reference the center of gravity consists of its $X,Y$ coordinates and its heading $\psi$. The parameters and inputs of the model are the same as of the kinematic bicycle model presented in Table \ref{bicycleParameters.tab} in addition to the slip angles at the front and rear wheels $\alpha_f$ and $\alpha_r$. The state evolution of the system is defined as follows:

\begin{subequations}\label{EBM.eq}
    \begin{eqnarray}
        \dot X &=& V\cos{(\beta+\psi)} \label{vx_EBM.eq}\\
        \dot Y &=& V\sin{(\beta+\psi)} \label{vy_EBM.eq}\\
        \dot\psi &=& \frac{V\cos\beta(\tan(\delta-\alpha_f)+\tan\alpha_r)}{l_f+l_r} \label{psi_EBM.eq}\\
        \beta &=& \arctan\left(\frac{-l_f \tan\alpha_r + l_r \tan(\delta-\alpha_f)}{l_f+l_r} \right)\label{beta_EBM.eq}
    \end{eqnarray}
\end{subequations}
Although this model introduces accurate properties through the inclusion of the slip angles at the wheels, the difficulty remains in identifying the slip angles without referring to the vehicle's dynamics. The proposed approach solves this problem by using recurrent neural networks as we will discuss next. 

\section{The Hybrid Extended Bicycle Model (HEBM)}\label{proposedApproach.sec}
The proposed approach aims to provide a simple model able to accurately represent the behavior of the vehicle in high dynamics. The developed model will be used later on to plan vehicle trajectories. As mentioned previously, the extended bicycle model will be at the core of the proposed approach while integrating recurrent neural networks to predict the wheels' slip angles. The use of recurrent neural networks requires the creation of a training data set, this will be presented in Section \ref{dataSet.ssec}; then, the network architecture and the training details will be presented in Section \ref{architecture.ssec}. 

\subsection{Data Generation}\label{dataSet.ssec}
Introducing a slip angle predictor based on recurrent neural networks necessitates the creation of a training data set. As we mentioned earlier, the model presented in \cite{polack_consistency_2018} and detailed in Section \ref{referenceModel.sec} is used as our reference model; thus, we seek to develop a hybrid extended bicycle with a close behavior to the mentioned reference model. This implies the use of the reference model to generate the training samples. 

To generate the training samples specific torques and steering angles should be applied as control inputs to the simulator. The data set is made of 5000 2-second trajectories sampled at 100 Hz (a total of 1 million samples). The used procedure to create the data set is the following:
\begin{enumerate}
    \item A random initial vehicle state is chosen.  
    \item Random controls are drawn from a uniform distribution as follows:
    \begin{itemize}
        \item A torque distribution between $\SI{0}{\newton\meter}$ and $\SI{800}{\newton\meter}$ if $V<\SI{10}{\meter\per\second}$, $\SI{-1000}{\newton\meter}$ and $\SI{800}{\newton\meter}$ if $\SI{10}{\meter\per\second}<V<\SI{30}{\meter\per\second}$ and $\SI{-1000}{\newton\meter}$ and $\SI{0}{\newton\meter}$ if $V>\SI{30}{\meter\per\second}$.
        \item A steering angle distribution between $\SI{-0.5}{\radian}$ and $\SI{0.5}{\radian}$. 
    \end{itemize}
    \item The sampled controls are applied to the 9 DoF model and are held constant for a uniformly drawn period between $\SI{0.01}{\second}$ and $\SI{1}{\second}$. The state of the reference model is updated accordingly. 
    \item The procedure repeats from Step (2) until a 2-second trajectory is formed. 
\end{enumerate}
The created data set is used to train the network defined next. 

\begin{figure*}[t]
    \vspace{.1cm}
    \centering
    \includegraphics[width=.9\textwidth]{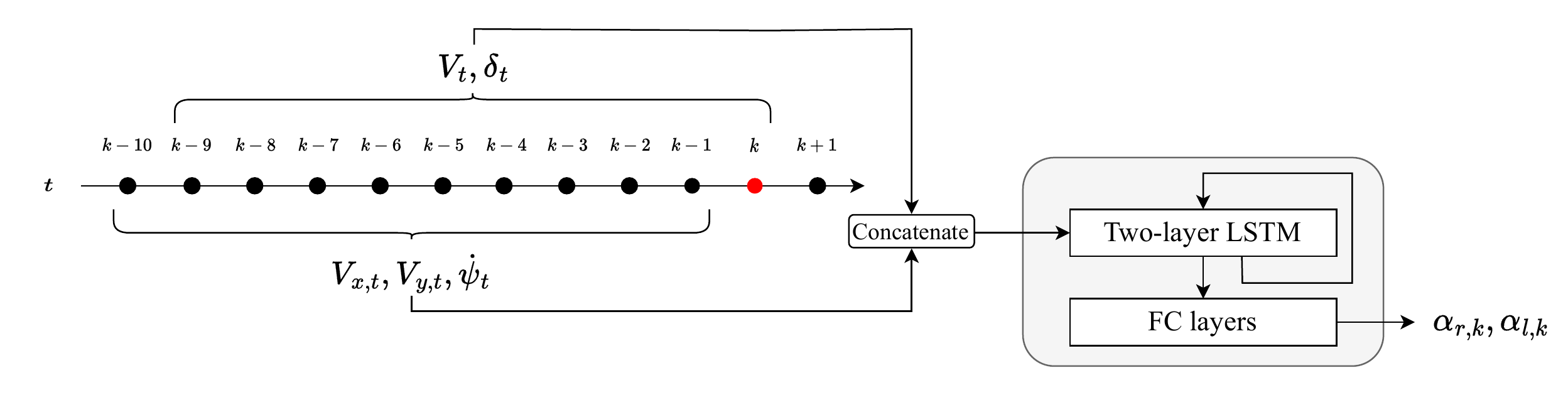}
    \caption{Wheel slip predictor architecture.}
    \label{architecture.fig}
\end{figure*}

\subsection{Slip Angle Predictor}\label{architecture.ssec}
The previously presented extended bicycle model lacks the knowledge of the slip angles at the front and rear wheels to be able to operate. In other words, the state evolution of the model described by Equations (\ref{EBM.eq}) requires the values of the slips at the wheels to compute the next state.  We propose to use recurrent neural networks (RNNs) to access these quantities. This choice is motivated by the ability of recurrent neural networks to perform well with dynamical systems \cite{graber_hybrid_2019}. The proposed predictor involves Long Short-Term Memory (LSTM) networks, a variant of RNNs. 

The proposed architecture is shown in Fig. \ref{architecture.fig}. As shown in the figure, to be able to predict the slips at $t=k$ the inputs to the network are split between the vehicle state part that involves the longitudinal and lateral velocities and yaw rate for the last 10 time steps ($t=k-1..k-10$), and the controls part that involves the controls ($V$, $\delta$) applied at the last 9 time steps and the ones to be applied to reach the state at the current time step ($t=k..k-9$). The outputs of the network are the resulting slip angles at the wheels at $t=k$ when the considered controls are applied. The output of the network will allow the computation of the state evolution of the model from $t=k-1$ to $t=k$ using Equations (\ref{EBM.eq}).

In brief, at each time step the controls to be applied and the current state of the model are used in addition to the slip angles predicted by the network to compute the next state of the model. 

The network is made of two LSTM layers consisting of 32 and 64 neurons respectively and three fully connected layers of 128, 256 and 128 neurons respectively. ReLU activation functions are used. 

The loss function employed consists of using the predicted slips to compute the state evolution of the model and comparing the resulting longitudinal and lateral velocities and yaw rate to the reference data. Thus, after each forward pass through the network, having the outputs $\alpha_{f}, \alpha_{r}$, the side-slip angle is calculated using Equation (\ref{beta_EBM.eq}) and the state evolution of the extended bicycle model is calculated using Equations (\ref{vx_EBM.eq})-(\ref{psi_EBM.eq}), then the velocities are transformed to the vehicle frame using the following equations:
\begin{subequations}
    \begin{eqnarray}
        V_x &=& \dot Y \sin\psi + \dot X \cos\psi \\
        V_y &=& \dot Y \cos\psi - \dot X \sin\psi
    \end{eqnarray}
\end{subequations}
These velocities along with the yaw rate are compared to the reference through an L1 loss that will be back propagated through the network. The loss function used will emphasize on lateral dynamics as we have noticed that learning the longitudinal dynamics was easier for the network. Hence, the loss function used is defined as follows:
\begin{equation}
    L = 0.2L_1^{V_x}+0.4L_1^{V_y}+0.4\frac{1}{\gamma}L_1^{\dot\psi}
\end{equation}
with $\gamma=0.05$ being a scaling factor to account for the scaling difference between $V_x$, $V_y$ and $\dot\psi$.

The network is implemented using PyTorch and is trained using the Adam optimizer with a batch size equal to $64$ and an initial learning rate equal to $1e-04$.

The trained slip predictor will be used, inside an extended bicycle model as described previously, resulting in the hybrid extended bicycle model. The resulting model will be employed in a planner next. 

\section{Planner}\label{planner.sec}
The aim of this work is to create a model able to accurately represent the vehicle's behavior, to be used for planning feasible trajectories even in high dynamics. The hybrid extended model developed in the previous section will be used for this purpose. Given that the developed model involves complexities due to the used RNNs, implementing it in a classical optimal control application would be complicated and computationally demanding. For this reason, a Model Predictive Path Integral (MPPI) based planner is discussed. A review of the MPPI technique will be presented in Section \ref{mppiReview.ssec}; the MPPI technique will be applied to plan feasible trajectories using the developed model in Section \ref{mppi-ebm-Planner.ssec}; low level controllers that will apply controls to the simulator to follow the planned trajectories are presented in Section \ref{lowLevelControl.ssec}.

\subsection{Review of MPPI}\label{mppiReview.ssec}
The Model Predictive Path Integral (MPPI) \cite{williams_aggressive_2016} is a sampling-based, derivative-free, model predictive control algorithm. It consists of using 
a Graphics Processing Unit (GPU) to sample a large number of trajectories based on a specific model at each time step which evaluation will lead to the computation of the optimal control. 

Algorithm \ref{mppi.alg} describes the operation of the MPPI method. The algorithm starts by defining the number of samples, which is the number of trajectories $\tau$ to be generated, each having $N$ time steps. The different trajectories are generated based on a model described by the function $f$ to which inputs $u+\delta u$ are applied. $Z_{t_0}$ denotes the initial state of the model. $\delta u$ is sampled from a uniform distribution and added to an initial control sequence updated at each iteration. The term $S$ represents the cost function that involves a running cost term $\hat{q}$ computed at each time step and a terminal cost term $\phi$ computed at the end of the generation process, $\hat{q}$ being:
\begin{equation}
    \hat{q} = q(Z) + \frac{1-\nu ^{-1}}{2}\delta u^T R \delta u + u^T R \delta u + \frac{1}{2} u^T R u
\end{equation}
$\nu$ being the exploration noise which determines how aggressively MPPI explores the state space and R being a positive definite control weight matrix. 
The control sequence is then updated according to the cost of each trajectory while taking into consideration the minimum cost and the inverse temperature $\lambda$ that impacts the degree of selectiveness of the algorithm. The updated control sequence is smoothed using the Savitzky-Golay convolutional filter. The first control of the sequence is returned and the process reiterates.

\begin{algorithm}
\caption{MPPI Algorithm}\label{mppi.alg}
\begin{algorithmic}
\State \textbf{Given:} $K:$ Number of Samples;\\
$N:$ Number of time steps;\\
$(u_0,u_1,...u_{N-1})$: Initial control sequence;\\
$f, \Delta T$: Dynamics function, step size;\\
$\phi, q, \lambda$: Cost function/Hyperparameters;\\
SGF: Savitzky-Golay (SG) convolutional filter;

\While{\textit{task not completed}}
    \State $\delta u \gets $ RandomNoiseGenerator();
    \State $\hat{S}(\tau_k) \gets$ CostInitializer();
    \For {$k \gets 0$ \textbf{to} $K-1$}
        \State $Z \gets Z_{t_0}$;
        \For{$t \gets 1$ \textbf{to} $N$}
            \State $Z_{t+1} \gets Z_t + f(Z_t, u_t+\delta u_{t,k}) \Delta T$;
            \State $\hat{S}(\tau_{t+1,k}) \gets \hat{S}(\tau_{t,k}) + \hat{q}$;
        \EndFor
        \State $\hat{S}(\tau_k) \gets \hat{S}(\tau_{N,k}) + \phi (z_N)$;
    \EndFor
    \State $\hat{S}_{\text{min}} \gets \text{min}_k[\hat{S}(\tau_k)]$;
    \For{$t \gets 0$ \textbf{to} $N-1$}
        \State $u_t \gets u_t + \frac{\sum_{k=1}^{K} \exp \left(-(1/\lambda)[\hat{S}(\tau_{k})-\hat{S}_{\text{min}}]\right)\delta u_{t,k}}{\sum_{k=1}^{K} \exp \left(-(1/\lambda)[\hat{S}(\tau_{k})-\hat{S}_{\text{min}}]\right)}$;
    \EndFor
    \State $u \gets \text{SGF}(u)$;
    \State Apply($u_0$);
    \For{$t \gets 1$ \textbf{to} $N-1$}
        \State $u_{t-1} \gets u_t$;
    \EndFor
    \State $u_{N-1} \gets \text{Initialize}(u_{N-1})$;
\EndWhile
\end{algorithmic}
\end{algorithm}

The presented algorithm will be used for planning trajectories using the previously presented hybrid model. The implementation details are discussed next. 

\subsection{MPPI-based Hybrid EBM Planner}\label{mppi-ebm-Planner.ssec}
\begin{figure}
    \centering
    \includegraphics[width=\columnwidth]{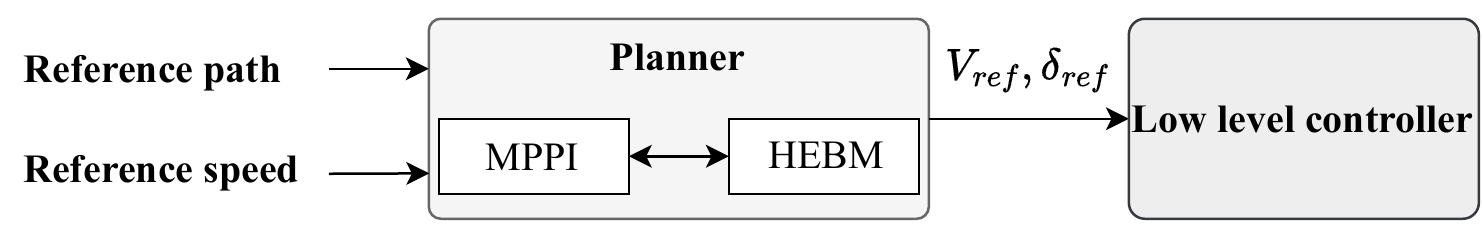}
    \caption{Planner-Controller proposed architecture.}
    \label{plan-control.fig}
\end{figure}

The proposed planner will be based on an MPPI approach with a hybrid extended bicycle model. The planner will make use of Algorithm \ref{mppi.alg} detailed in the previous section. In the proposed planner, the function $f$ mentioned above represents the state evolution presented in Equations (\ref{EBM.eq}). The architecture of the planner is shown in Fig. \ref{plan-control.fig}. It takes as inputs the reference path and speed to compute optimal trajectories and output to the low level controller the velocity and steering angle to follow for a minimal cost. The used running cost associated with the MPPI planner is defined as:
\begin{equation}
\begin{split}
    q(Z) = (Z-Z^{ref})^TQ_Z(Z-Z^{ref}) + \\ (V-V^{ref})^TQ_V(V-V^{ref})
\end{split}
\end{equation}
Knowing that the state $Z$ of the extended bicycle model is $Z=\begin{bmatrix}
    X & Y & \psi
\end{bmatrix}$ which is compared to the reference path's $X$ and $Y$ coordinates and heading. The used terminal cost is $\phi(Z_T) = q(Z_T)$. The parameters defining the used MPPI method are shown in Table \ref{mppiParameters.tab}. The random noise values $\delta u$ are issued from a normal distribution between $\SI{-0.08}{\meter\per\second}$ and $\SI{0.05}{\meter\per\second}$ for the velocity $V$ control and between $\SI{-0.02}{\radian}$ and $\SI{0.02}{\radian}$ for the steering angle $\delta$ control. The cost matrices are: $Q_Z = 4\text{ . Diag}(1,1,10)$, $Q_V = 3$, $R =  1e-02 \text{ . Diag}(1,1)$.

At each iteration the proposed planner will use the current reference vehicle state as a starting point to sample $K=1024$ trajectories for $N=100$ time steps using the hybrid extended model that takes as control inputs $V$ and $\delta$; the MPPI algorithm is then used to compute the optimal control sequence. The first five computed controls are sent to the low level controllers to actuate the reference vehicle accordingly. The planner will run at $\SI{20}{\hertz}$ while the low level controllers introduced next will run at $\SI{100}{\hertz}$.

\begin{table}
\vspace{0.2cm}
\centering
\caption{MPPI Parameters}
    \begin{tabular}{c c}
      \toprule
      \textbf{Parameter} & \textbf{Value} \\
      \midrule
      $\lambda$ & $3e-01$ \\
      $\nu$ & $1000$ \\
      $K$ & 1024 \\
      $N$ & 100 \\
      $\Delta T$ & \SI{0.01}{\second} \\
      \bottomrule
    \end{tabular}
\label{mppiParameters.tab}
\end{table}

\subsection{Low level controllers}\label{lowLevelControl.ssec}
In the architecture proposed above, the MPPI planner would generate reference velocities and steering for the low level controllers to follow. The low level controllers will apply torques and steering to the reference four-wheel model introduced previously. The low level control is split into longitudinal control (torques) and lateral control (steering angle); each will be treated separately. 

The longitudinal controller aims to make the reference vehicle's velocity reach the reference velocity received from the planner; thus, the error it is trying to minimize is defined as $e_V = V_{\text{ref}}-V_{\text{vehicle}}$ with $V_{\text{vehicle}}$ being the current simulator velocity computed from the state introduced in Section  \ref{referenceModel.sec} using $V = \sqrt{V_x^2 + V_y^2}$. A PID controller, with gains $K_{P,V}$, $K_{D,V}$ and $K_{I,V}$ is used. 

The lateral controller follows the approach introduced in \cite{polack_consistency_2018}. The lateral control is split to two parts: an open loop part and a closed loop part. The open loop part applies the reference steering angle received by the planner; the closed loop part consists of a PID controller that aims to reduce the heading error projected 5 time steps ahead by applying the reference velocity and steering angle to an HEBM. The gains are defined as $K_{P,\delta}$, $K_{D,\delta}$, and $K_{I,\delta}$.

The used gains for both longitudinal and lateral PID are shown in Table \ref{pidParameters.tab}. The proposed approach is evaluated next.

\begin{table}
\vspace{0.2cm}
\centering
\caption{Longitudinal and lateral PID Parameters}
    \begin{tabular}{c c c c c c}
      \toprule
      \textbf{$K_{P,V}$} & \textbf{$K_{D,V}$} & \textbf{$K_{I,V}$}& \textbf{ $K_{P,\delta}$}& \textbf{ $K_{D,\delta}$}& \textbf{ $K_{I,\delta}$}\\
      \midrule
      $500$ & $20$ & $15$ & $0.2$ & $0.01$ & $0.005$ \\
      \bottomrule
    \end{tabular}
\label{pidParameters.tab}
\end{table}

\section{Results}\label{results.sec}
The proposed planning architecture is evaluated in this Section. The aim is to control the four-wheel vehicle presented in \ref{referenceModel.sec} based on the approach presented above. As mentioned previously, the planner will get as inputs the reference path and a reference speed which will vary between test cases; will compute optimal trajectories and return reference velocities and steering angles for the low level controllers. The low level controllers will apply the torque and steering controls accordingly to the four-wheel vehicle model. We compare our approach to a kinematic bicycle model (KBM) based approach for the reasons stated previously; for the sake of fairness, we implement the same architecture by replacing the HEBM in Fig. \ref{architecture.fig} by the KBM. 

In the following we start by validating our method based on the oval trajectory validation tests used in \cite{williams_information-theoretic_2018} and \cite{spielberg_neural_2022}. We then move to a lane change maneuver validation to assess the performance of our method in real life scenarios. 

\subsection{Oval trajectory testing}

\begin{figure}
    \centering
    \includegraphics[width=\columnwidth]{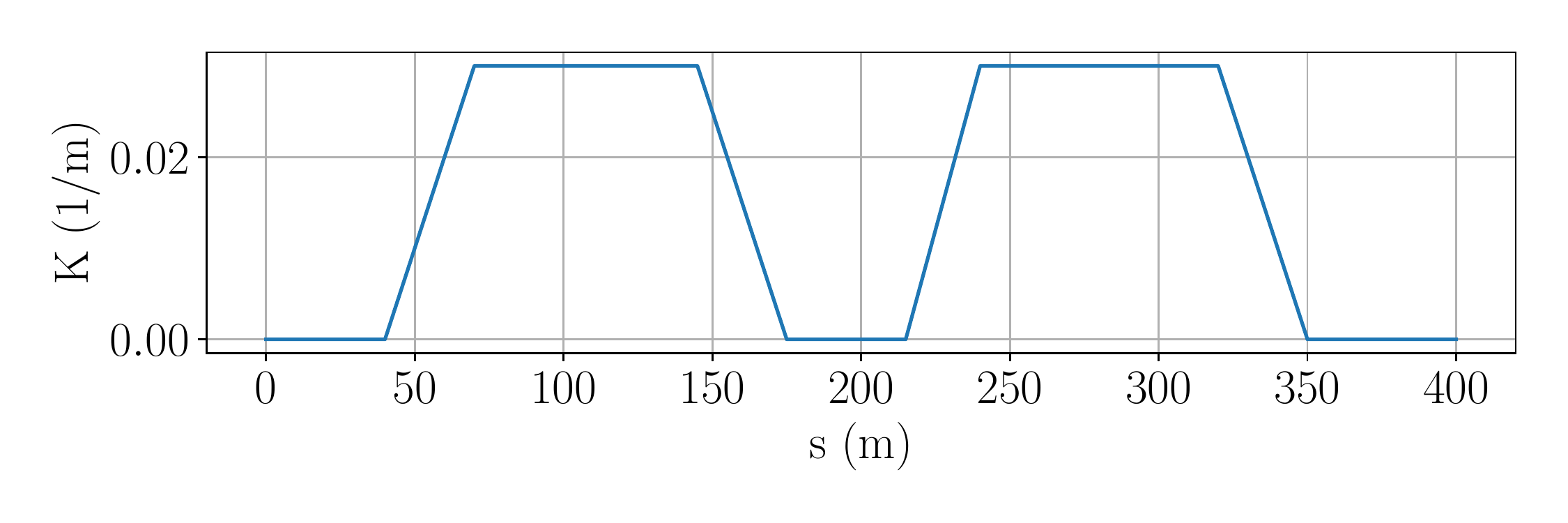}
    \caption{Curvature of the reference oval path.}
    \label{pathCurvature.fig}
\end{figure}

The test consists of running the vehicle on an oval trajectory with different speeds, clockwise (CW) and counterclockwise (CCW). The curvature of the reference path is shown in Fig. \ref{pathCurvature.fig}. To be able to compare the two methods at different speeds, we present in Table \ref{ovalResults.tab} the different tests while showing the mean absolute lateral error (MAE), the max absolute lateral error and the mean velocity associated with each test. The lateral error being defined as:
\begin{equation}
    \text{error} = (Y_{\text{vehicle}}-Y_{\text{ref}})\cos\psi_{\text{ref}} - (X_{\text{vehicle}}-X_{\text{ref}})\sin\psi_{\text{ref}}
\end{equation}
$X_{\text{ref}}$, $Y_{\text{ref}}$, $\psi_{\text{ref}}$ being the closest reference points to the vehicle. 
We remark that for both models higher velocities are associated with higher errors. Though, the KBM based planner is not comfortable with high velocities as it is not able to stick with the reference speed and have minimal errors simultaneously. On the other hand, the proposed HEBM approach is able to perform maneuvers with higher velocities, sticking with the desired velocity of each case, while maintaining low lateral errors. It can be seen e.g. that for a $V_{\text{desired}} = \SI{18}{\meter\per\second}$ (CCW), the HEBM is able to reach high velocities, performing maneuvers with $a_y^{\text{max}}=0.76g$ while keeping a maximum lateral error of $\SI{0.88}{\meter}$ while the KBM approach can't keep with the desired velocity and shows higher errors. The HEBM planner has lower errors and higher velocities in all of the tested scenarios.

\begin{figure}[htbp]

\subfloat[Comparison between the vehicle behavior when using the KBM-based and HEBM-based planners for the CW scenario at $V_{\text{desired}}=\SI{18}{\meter\per\second}$]{%
  \includegraphics[width=\columnwidth]{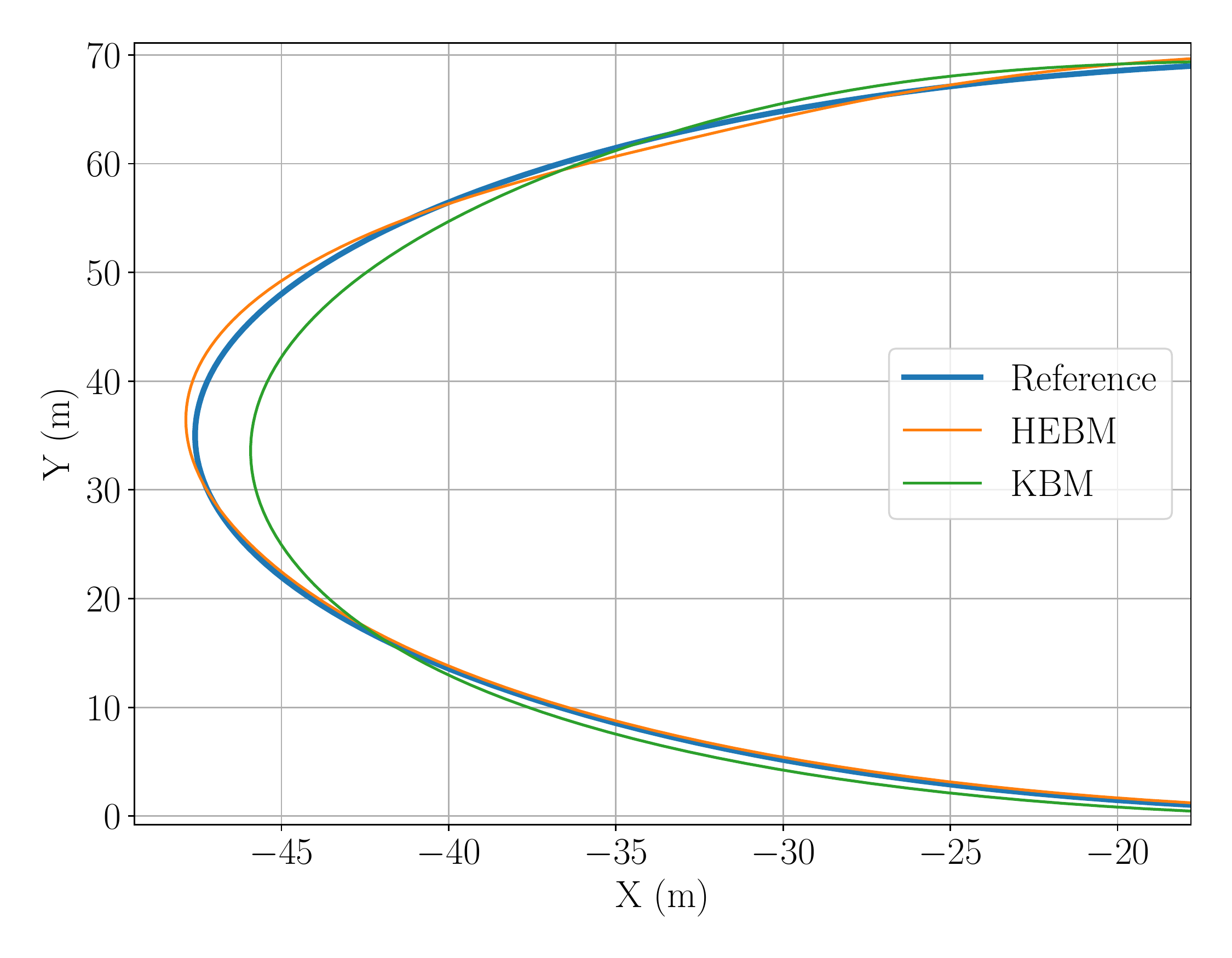}%
}
\hfill
\subfloat[Comparison between the vehicle behavior when using the KBM-based and HEBM-based planners for the CCW scenario at $V_{\text{desired}}=\SI{18}{\meter\per\second}$]{%
  \includegraphics[width=\columnwidth]{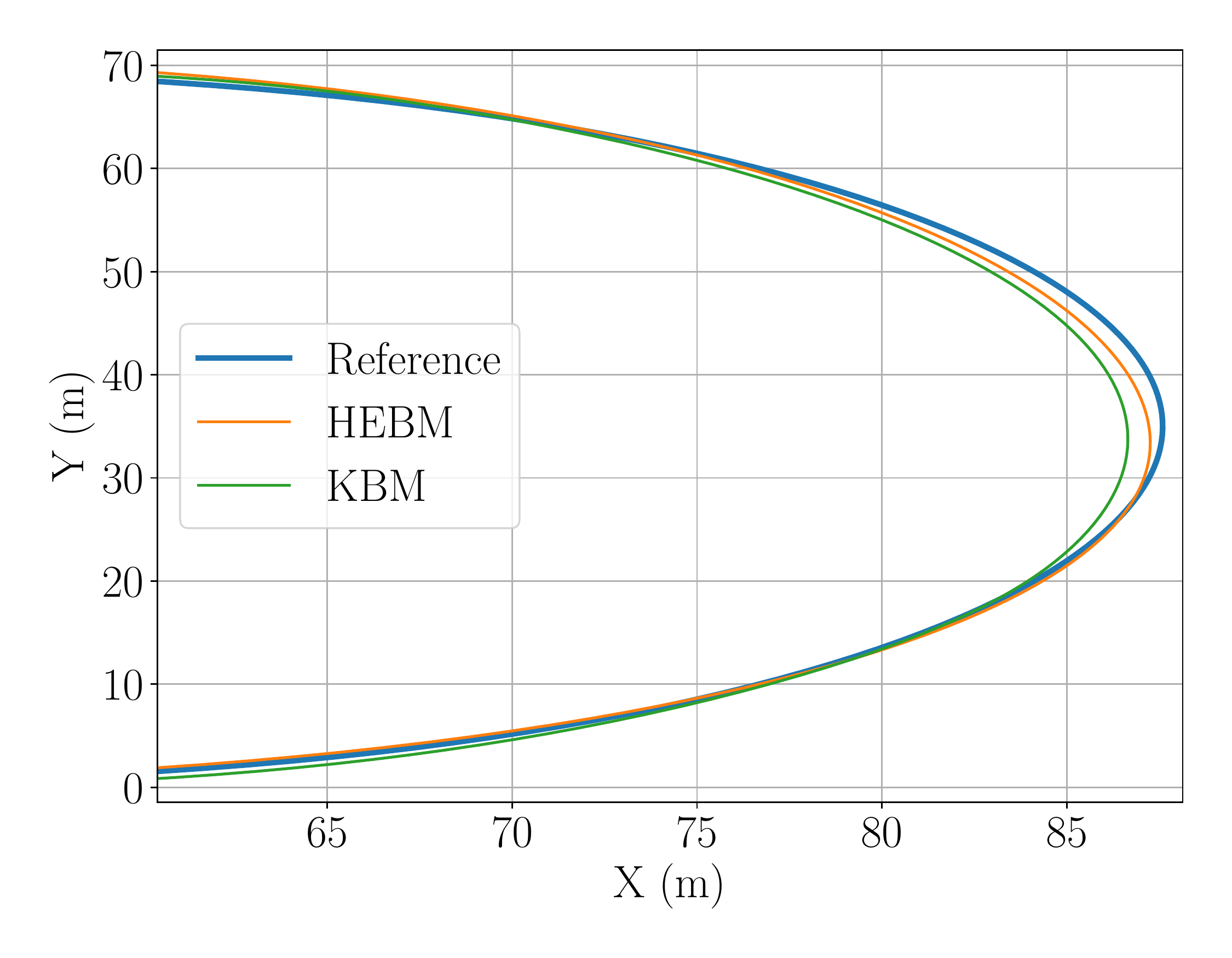}%
}
\caption{Comparison between the behavior of the vehicle subject to the KBM and HEBM planners at their highest lateral error point. The proposed approach is able to be closer to the reference path while maintaining higher velocities as shown in Table \ref{ovalResults.tab}.\label{ovalError.fig}}

\end{figure}

A close-up on the behavior of the vehicle under the two planners at their highest lateral error points is shown in Fig. \ref{ovalError.fig}. The plots show that the proposed planner is able to make the vehicle drive closer to the reference path.

\begin{table}
\vspace{0.2cm}
\centering
\caption{Oval trajectory testing results showing the ability of the proposed HEBM method to present lower lateral errors and to drive with higher velocities. (MAE and Max Error in $\SI{}{\meter}$, $V_{\text{desired}}$ and mean $V$ in $\SI{}{\meter\per\second}$).}
    \begin{tabular}{c c c c c c}
      \toprule
      \textbf{$V_{\text{desired}}$} & \textbf{Method}& \textbf{Direction} & \textbf{MAE} & \textbf{Max Err.} & \textbf{Mean $V$} \\
      \midrule
      $8$ & HEBM & CW & \textbf{0.05} & 0.26 & 7.97\\
      $8$ & KBM & CW & 0.08 & 0.26 & 7.16\\\midrule
      $10$ & HEBM & CW & \textbf{0.072} & \textbf{0.29} & 10.2\\
      $10$ & KBM & CW & 0.13 & 0.59 & 8.33\\\midrule
      $12$ & HEBM & CW & \textbf{0.09} & \textbf{0.31} & 12.56\\
      $12$ & KBM & CW & 0.18 & 0.66 & 9.18\\\midrule
      $14$ & HEBM & CW & \textbf{0.16 }& \textbf{0.48} & 14.7\\
      $14$ & KBM & CW & 0.17 & 0.66 & 9.11\\\midrule
      $16$ & HEBM & CW & \textbf{0.11} & \textbf{0.56} & 15.86\\
      $16$ & KBM & CW & 0.17 & 0.81 & 9.44\\\midrule
      $18$ & HEBM & CW & \textbf{0.2} &\textbf{ 0.67} & 16.4\\
      $18$ & KBM & CW & 0.33 & 1.77 & 10.85\\
      \midrule
      \midrule
      $8$ & HEBM & CCW & \textbf{0.05} & \textbf{0.22} & 7.97\\
      $8$ & KBM & CCW & 0.1 & 0.38 & 7.6\\\midrule
      $10$ & HEBM & CCW & \textbf{0.07} & \textbf{0.29} & 10.2\\
      $10$ & KBM & CCW & 0.12 & 0.44 & 7.9\\\midrule
      $12$ & HEBM & CCW & \textbf{0.07} & \textbf{0.4} & 12.25\\
      $12$ & KBM & CCW & 0.14 & 0.57 & 8.72\\\midrule
      $14$ & HEBM & CCW & \textbf{0.08 }& \textbf{0.45} & 14.32\\
      $14$ & KBM & CCW & 0.2 & 0.69 & 9.91\\\midrule
      $16$ & HEBM & CCW & \textbf{0.15} & \textbf{0.75} & 16.2\\
      $16$ & KBM & CCW & 0.2 & 0.81 & 10.1\\\midrule
      $18$ & HEBM & CCW & \textbf{0.2} &\textbf{0.88} & 17.8\\
      $18$ & KBM & CCW & 0.3 & 1.11 & 11.5\\
      \bottomrule
    \end{tabular}
\label{ovalResults.tab}
\end{table}

\subsection{Lane change testing}
\begin{figure}
    \centering
    \includegraphics[width=\columnwidth]{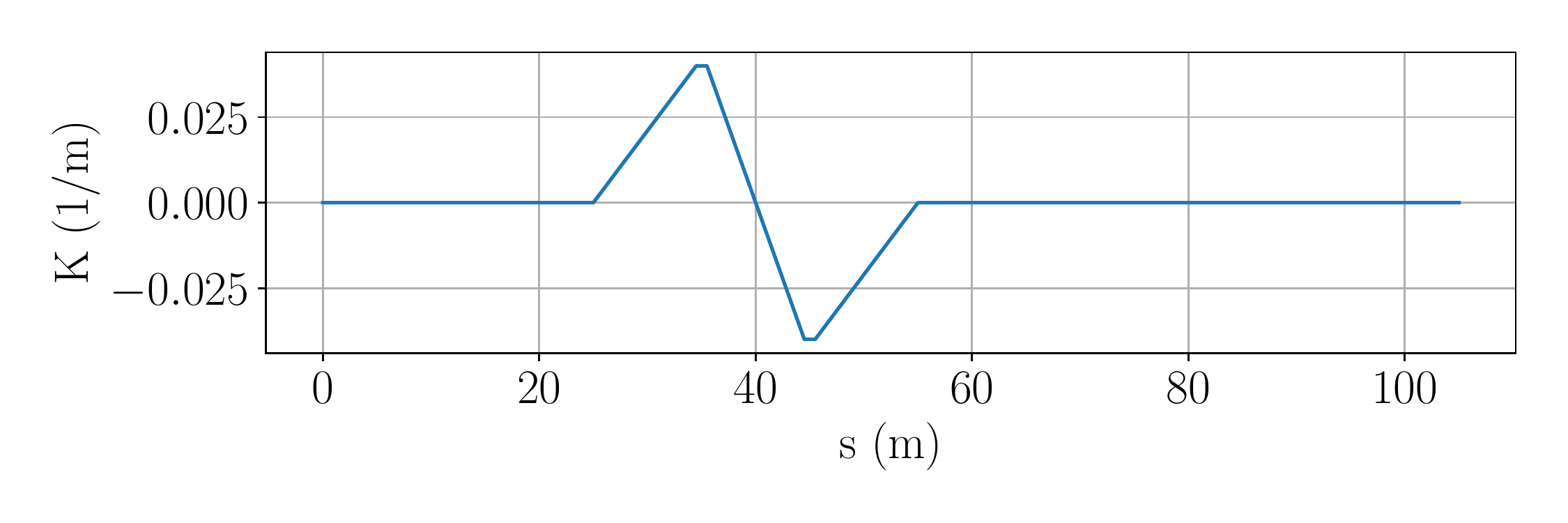}
    \caption{Curvature of the reference lane change path.}
    \label{pathCurvature2.fig}
\end{figure}

To further validate our method, we test its behavior when effecting a lane change maneuver. The lane change maneuver is based on the ISO 3888-1 standard and its curvature is shown in Fig. \ref{pathCurvature2.fig}. Similarly to the previous test, we assess the performance of both approaches on different speeds. The results are seen in Table \ref{lcResults.tab}. The table shows that the KBM based planner is able to drive the vehicle into higher velocities as compared with the previous test, this is due to the relaxed heading constraints of the lane change maneuver as opposed to the oval maneuver. The behavior of the KBM based planner lacks accuracy. The HEBM based planner is able to stick to the reference path with lower errors even at high velocities, while the KBM based planner loses accuracy with higher velocities. An illustration of the highest error case ($V_{\text{desired}}=\SI{25}{\meter\per\second}$) is seen in Fig. \ref{laneChange.fig}. The figure shows the behavior of both planners. The lateral acceleration of the vehicle under the HEBM planner reaches $a_y^{\text{max}}=0.75g$. The KBM planner is not able to effect the maneuver accurately, thus safely. 

In brief, the proposed approach is able to accurately follow the reference path provided to the planner while keeping with the reference velocity, while the kinematic bicycle model fails to follow the provided path especially at higher velocities i.e. higher lateral accelerations. 

\begin{table}
\vspace{0.2cm}
\centering
\caption{Lane change trajectory testing results showing that the HEBM method presents lower lateral and velocity errors. (MAE and Max Error in $\SI{}{\meter}$, $V_{\text{desired}}$ and mean $V$ in $\SI{}{\meter\per\second}$).}
    \begin{tabular}{c c c c c}
      \toprule
      \textbf{$V_{\text{desired}}$} & \textbf{Method} & \textbf{MAE} & \textbf{Max Err.} & \textbf{Mean $V$} \\
      \midrule
      $10$ & HEBM & \textbf{0.04} & \textbf{0.2} & 10.1\\
      $10$ & KBM & 0.09 & 0.4 & 8.47\\\midrule
      $15$ & HEBM & \textbf{0.05} & \textbf{0.27} & 15.02\\
      $15$ & KBM & 0.17 & 0.74 & 12.35\\\midrule
      $20$ & HEBM & \textbf{0.11} & \textbf{0.47} & 20\\
      $20$ & KBM & 0.38 & 1.12 & 17.93\\\midrule
      $25$ & HEBM & \textbf{0.2} & \textbf{0.51} & 25.13\\
      $25$ & KBM & 0.75 & 3.7 & 27.2\\
      \bottomrule
    \end{tabular}
\label{lcResults.tab}
\end{table}

\section{Conclusion}\label{conclusion.sec}
In this paper, we proposed a simple model that combines an extended bicycle with LSTMs to model the state of the vehicle in challenging maneuvers. The aim was to use the proposed model for planning purposes. 

We started by defining our reference vehicle and then presented two simplified vehicle models: the kinematic bicycle model and the extended bicycle model. The KBM was introduced for comparison purposes given its popularity in the literature while the EBM was introduced as the basis of the proposed approach. 

We augmented the EBM using LSTMs to be able to predict the slip angles at the wheels. The augmented model was implemented in an MPPI-based planning approach. The proposed planner computes optimal trajectories and outputs reference velocities and steering to low level controllers that control the four-wheel vehicle's torque and steering. 

Two validation tests were performed, the oval maneuver and the lane change maneuver. The proposed approach was compared to a kinematic bicycle based approach. The performed tests showed that our approach is able to drive the vehicle in a more accurate way while keeping high velocities while the other approach failed to deliver accurate behavior. The proposed approach is able to cope with both low and high dynamic maneuvers making a vehicle using it able to navigate safely despite the harshness of the scenario. 

Future work would apply this method to real vehicles and explore its limitations. 

\bibliographystyle{ieeetr}
\bibliography{main}

\end{document}